\documentclass[runningheads]{llncs}
\usepackage{accv}
\usepackage{accvabbrv}
\usepackage{multirow}
\usepackage{adjustbox}

\usepackage{graphicx}
\usepackage{booktabs}
\usepackage{tabularx} 
\usepackage{booktabs} 
\usepackage{multirow} 
\usepackage{array}    
\usepackage{caption}  
\usepackage{amsmath} 
\usepackage{algorithm}
\usepackage{algorithmic}
\usepackage{hyperref}
\usepackage{wrapfig}
\usepackage{graphicx}
\usepackage{lipsum}  
\usepackage{wrapfig}
\usepackage{algorithm}
\usepackage{colortbl}
\usepackage{algorithmic}
\usepackage{graphicx}
\usepackage{subcaption}
\usepackage{xcolor}
\usepackage{float}
\usepackage{amsmath}
\usepackage{listings}
\usepackage{bbm}
\usepackage{amsmath}
\usepackage{booktabs}
\usepackage{amssymb}
\usepackage{pgfplots} 
\usepackage{tikz} 
\usepackage{graphicx} 
\usepackage{caption} 
\usepackage{subcaption} 
\pgfplotsset{compat=1.17} 
\usepgfplotslibrary{groupplots} 
\usetikzlibrary{patterns} 
\usepackage[accsupp]{axessibility}  
\usepackage{hyperref}
\usepackage{orcidlink}

\begin{document}

\title{CoVLM: Leveraging Consensus from Vision-Language Models for Semi-supervised Multi-modal Fake News Detection}
\titlerunning{CoVLM for Semi-supervised MFND}

\author{Devank \and Jayateja Kalla \and Soma Biswas}

\authorrunning{Devank et al.}

\institute{Department of Electrical Engineering, \\Indian Institute of Science, Bangalore, India. \\
\email{\{devank2022, jayatejak, somabiswas\}@iisc.ac.in}}

\maketitle

\begin{abstract}
In this work, we address the real-world, challenging task of out-of-context misinformation detection, where a real image is paired with an incorrect caption for creating fake news. Existing approaches for this task assume the availability of large amounts of labeled data, which is often impractical in real-world, since it requires extensive manual intervention and domain expertise. In contrast, since obtaining a large corpus of unlabeled image-text pairs is much easier, here, we propose a semi-supervised protocol, where the model has access to a limited number of labeled image-text pairs and a large corpus of unlabeled pairs. Additionally, the occurrence of fake news being much lesser compared to the real ones, the datasets tend to be highly imbalanced, thus making the task even more challenging. Towards this goal, we propose a novel framework, \textbf{Co}nsensus from \textbf{V}ision-\textbf{L}anguage \textbf{M}odels (CoVLM) \footnote{Code Link \url{https://github.com/devank3/CoVLM}}, which generates robust pseudo-labels for unlabeled pairs using thresholds derived from the labeled data. This approach can automatically determine the right threshold parameters of the model for selecting the confident pseudo-labels. Experimental results on benchmark datasets across challenging conditions and comparisons with state-of-the-art approaches demonstrate the effectiveness of our framework.
 
\end{abstract}

\section{Introduction}

\label{sec:intro}

\begin{figure}[t]
    \centering
    \includegraphics[width=0.9\textwidth]{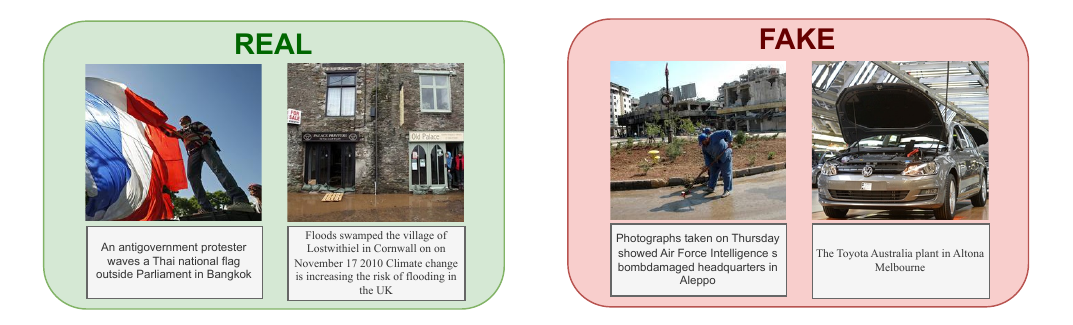}
    \caption{A sample real and fake image-pair from the NewsCLIPpings dataset~\cite{luo2021NewsCLIPpings}. The model needs to capture the subtle inconsistencies between the image and text pairs to understand their authenticity. }
    \label{fig:figure1}
\end{figure}

The proliferation of fake news in social media has made fake news detection a critical task for maintaining information integrity \cite{Omar2024FakeNews}, safeguarding public discourse \cite{shu2017fake}, and preventing the erosion of trust \cite{Zhou2020SurveyFakeNews}. 
One of the increasingly popular means of generating fake news is to pair real images with misleading/incorrect captions, since this requires minimal effort and technical expertise. 
Figure~\ref{fig:figure1} demonstrates few examples of real and fake image-text pairs from the benchmark NewsCLIPings Dataset~\cite{luo2021NewsCLIPpings}. 
Fake news often exhibit discrepancies between the visual content and the accompanying text, whereas real news tends to have a coherent relationship between images and text. 
Identifying these subtle inconsistencies can help to determine whether a given image-text pair is real or fake. Thus the goal of the existing out-of-context misinformation detection or multi-modal fake news detection (MFND) frameworks is to analyze large amounts of training data to learn these inconsistencies, which are used to infer whether a given test image-text pair is real or fake. 

One of the significant advancements in the direction of multi-modal machine learning~\cite{baltruvsaitis2018multimodal_multimodal_survey1, bayoudh2022survey_multimodal_survey2, zong2023self_multimodal_survey3} is to learn joint representations of image content and natural language. For example, the CLIP (Contrastive Language-Image Pre-training) model~\cite{radford2021learning} bridges the gap between image and natural language modalities by being trained on a huge dataset of image-caption pairs. Approaches leveraging these models for MFND task~\cite{luo2021NewsCLIPpings, abdelnabi2022open, yuan2023support, shalabi2024ooc, qi2024sniffer, liu2024mmfakebench, zhou2022fndclip, jiang2023sample} have shown promise. However, these approaches rely entirely on supervised data, i.e., image-text pairs labeled as real or fake. Annotating large amounts of data is extremely labor-intensive and requires domain expertise. For instance, verifying a news claim like {\em "An anti-government protester waves a Thai national flag
outside Parliament in Bangkok"} as demonstrated in Figure ~\ref{fig:figure1} involves significant expertise in international affairs and requires professionals who are proficient in global political dynamics. Conversely, collecting image-text pairs without annotations is much simpler.
In this work, we propose a realistic and practical Semi-Supervised MFND (SS-MFND) protocol, where the model has access to a few labeled image-text pairs and a large number of unlabeled pairs. 

\begin{figure}[t!]
    \centering
    \includegraphics[scale=0.68]{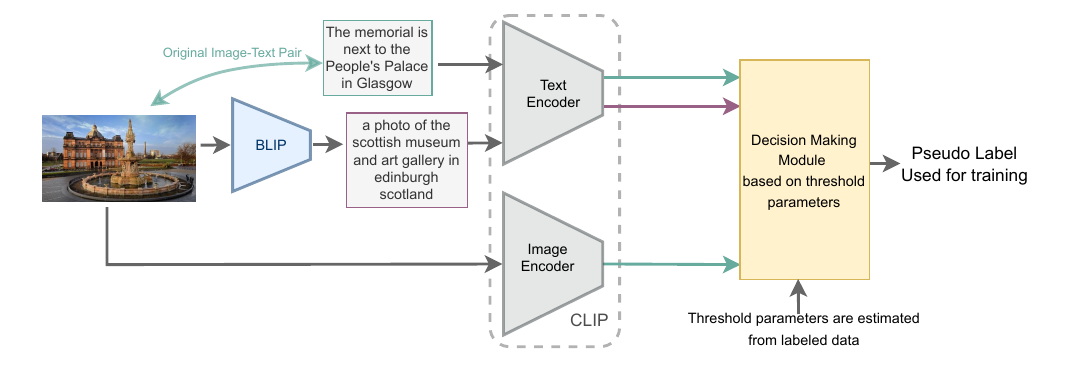} 
    \caption{Overview of CoVLM. For a given image-text pair, BLIP generates an additional image description. Using the original image, text, and the generated text, a decision is made on whether the pair is real or fake. This label is then used in training. The decision module’s threshold parameters are estimated from the labeled data.}
    \label{fig:intro_clip_blip}
\end{figure}

One of the most successful approaches to leverage unlabeled data in semi-supervised learning~\cite{sohn2020fixmatch} involves training a model first on the labeled data, followed by generating pseudo-labels for the unlabeled data and incorporating the confident ones into the training process. 
While such methods, like FixMatch \cite{sohn2020fixmatch}, Adsh~\cite{guo2022class}, and FreeMatch \cite{wang2022freematch}, have proven effective for unimodal data by utilizing the vast amounts of available unlabeled data, they fall short in semi-supervised MFND because they fail to capture the intricate relationships between real and fake image-text pairs. 
To address this challenge and generate robust pseudo-labels for unlabeled image-text pairs, we propose CoVLM, a novel approach that leverages the consensus between two vision-language models (VLMs), namely CLIP (Contrastive Language-Image Pre-training) \cite{radford2021learning} and BLIP (Bootstrapping Language-Image Pre-training) \cite{li2022blip}. 
The BLIP model generates descriptive text for a given image, which is then used in conjunction with the original text to generate robust pseudo-labels. This consensus from two models (CLIP, BLIP) ensures the robustness of the pseudo-labels. Figure~\ref{fig:intro_clip_blip} provides an overview of the CoVLM using CLIP and BLIP models.

Another challenge is that for image-text pairs collected from news-articles, blogs, posts, etc. the number of real pairs are far more compared to the number of fake pairs.
This results in severely class-imbalanced data, thereby making the task significantly more challenging. Since the benchmark NewsCLIPpings data is artificially created and balanced, it is not suited to analyze the performance of algorithms under a more realistic imbalanced scenario. 
Inspired by the rich literature on imbalanced semi-supervised learning for classification tasks \cite{hyun2021class, wei2021crest, oh2022daso, jiang2021sshr, arabmakki2016rls, zhang2020longtailed, fan2022coss, chen2021simple, guo2022class}, we introduce such imbalances in labeled and unlabeled data to evaluate the proposed approach under these realistic conditions. To this end, the contributions of this work can be summarized as follows:
\begin{enumerate}
    \item To the best of our knowledge, this is the first work to address the realistic and challenging Semi-Supervised multi-modal Fake News Detection task.
    \item We propose a novel framework CoVLM that utilizes vision-language models to generate robust pseudo-labels for the unlabeled image-text training pairs.
    \item Extensive experiments on widely used MFND datasets, namely NewsCLIPpings~\cite{luo2021NewsCLIPpings}, GossipCop~\cite{gossipcop}, and PolitiFact~\cite{politifact}, demonstrate the effectiveness of the proposed approach.
    \item In addition to the traditional balanced settings, we also test the framework in realistic imbalanced scenarios to evaluate its robustness.
\end{enumerate}
We now discuss the related work followed by the proposed framework and evaluation results.
\section{Related Work}
\label{relatedwork}
In this section we briefly discuss the related works on fake news detection and semi-supervised learning. \\ \\
{\bf Unimodal Fake News Detection: }Traditional approaches to fake news detection often focus on analyzing a single modality, such as text or image content in isolation. (i) Image Analysis: Prior research has explored image forensic features, semantic information, and statistical properties to detect manipulation~\cite{cao2020exploring_unimodal_survey_7}. Techniques for identifying image tampering can reveal signs of fake~\cite{chen2021image_unimodal_survey_8}. Additionally, common sense inconsistencies and poor image quality can be red flags for fake news~\cite{han2021fighting_unimodal_survey_15, li2021entity_unimodal_survey_26}. (ii) Text Analysis: Verifying logical consistency is crucial for detecting fake news in text format~\cite{conroy2015automatic_unimodal_survey_11}. Examining for grammatical errors, unusual writing styles, or specific rhetorical structures can also provide clues~\cite{conroy2015automatic_unimodal_survey_11, potthast2017stylometric_unimodal_survey_31}. However, both linguistic and visual patterns can be heavily influenced by the specific event and related domain knowledge. To address this challenge, Nan et al. ~\cite{nan2021mdfend_unimodal_survey_29} proposed utilizing a domain gate to combine representations learned from different experts, enabling their model to handle multi-domain fake news propagation within the text modality.
While these unimodal features offer valuable insights and play a significant role in distinguishing fake news, they neglect the crucial aspects of multi-modality - correlation and consistency between text and image content. This omission can hinder the overall effectiveness of these single modality methods when applied to multi-modal news. \\ \\
{\bf Multimodal Fake News Detection:} MFND focuses on utilizing both text and image modalities to detect fake news. 
Several approaches have been proposed to address this task, few notable ones being SAFE \cite{zhou2020safe}, Cultural Algorithm \cite{shah2020cultural}, Combination of textual, visual, and semantic information \cite{giachanou2020multimodal}, Fine-grained Classification \cite{segura2022multimodal}, FND-CLIP \cite{zhou2022fndclip}, ETMA \cite{yadav2022etma}, SAMPLE \cite{jiang2023sample}, DeBERTNeXT \cite{saha2023debertnext} and Tri Transformer-BLIP \cite{ttblip2024}.
Recently, researchers have also started gathering evidence from the internet to determine the authenticity of the image-text pair, and some notable works in this direction include NewsCLIPpings \cite{luo2021NewsCLIPpings}, Consistency-Checking Network \cite{abdelnabi2022open}, SEN \cite{yuan2023support}, OOCD \cite{shalabi2024ooc}, SNIFFER \cite{qi2024sniffer}, Zero-shot approach \cite{dey2024retrieval}, EVVER-Net \cite{chrysidis2024credible}, and MMFakeBench \cite{liu2024mmfakebench}.
But all these approaches assume that the entire training data is labeled, which is quite difficult for this task in realistic scenarios, because of the amount of human intervention required. 
Thus, we propose a real-world semi-supervised setting, which can utilize unlabeled data along with some labeled pairs for addressing this task. We also work in the closed setting, with no external evidence.  \\ \\
{\bf Semi-Supervised Learning: } In semi-supervised learning (SSL), various methods have been developed to effectively leverage unlabeled unimodal data to improve model performance. Numerous approaches have been proposed in the literature for the image domain \cite{sohn2020fixmatch, berthelot2019mixmatch, wang2022freematch, deng2022boosting, park2021opencos, kim2022conmatch, wallin2022doublematch, nguyen2023refixmatch, pham2021metapseudolabels, duan2022mutexmatch, li2021comatch, berthelot2019remixmatch}. FixMatch \cite{sohn2020fixmatch} is a significant work in this field, combining consistency regularization and pseudo-labeling with fixed thresholds. 
Building on the ideas of FixMatch, Adsh \cite{guo2022class} proposed using different thresholds for different classes to help underrepresented classes improve accuracy.
SSL in the text domain also has been extensively explored and continues to evolve with new methodologies and approaches \cite{Nigam2006, Dharmadhikari2012, Barman2020, Li2020, Li2019, Miyato2021, Duarte2023, Matsubara2024}. These works primarily focus on individual modalities either in the image or text domains within the context of SSL. However, our research aims to bridge these methodologies by concentrating on fake news detection involving both image and text components in a semi-supervised manner.

\section{Problem Definition}
We now formally define the problem of semi-supervised multi-modal fake news detection (SS-MFND), where the objective is to train a model  using a limited amount of labeled data and large amount of unlabeled training data, to determine whether a given image-text pair is real or fake at inference time. 
Specifically, the model has access to a labeled data $\mathcal{D}^{l} = \{(\mathcal{I}_i^l, \mathcal{T}_i^l, y_i^l)\}_{i=1}^{N^{l}}$ which consists of image-text pairs $(\mathcal{I}_i^l, \mathcal{T}_i^l)$ along with their corresponding labels $y_i^l$, indicating whether the pair is real or fake. Here, $N^{l}$ denotes the number of labeled data samples. Additionally, the model has access to an unlabeled dataset $\mathcal{D}^{ul} = \{(\mathcal{I}_i^{ul}, \mathcal{T}_i^{ul})\}_{i=1}^{N^{ul}}$, which contains image-text pairs without any information regarding their authenticity. 
It is usually easier to collect the image-text pairs compared to labeling them, which requires significant manual intervention and domain expertise. 
Notably, $N^{l} \ll N^{ul}$, indicating that the number of labeled samples is much smaller than the number of unlabeled samples in our experiments. 
We now discuss the proposed CoVLM framework in detail. 

\section{CoVLM for Semi-Supervised MFND}
\label{approach}
In our work, inspired by the recent literature in MFND~\cite{luo2021NewsCLIPpings, abdelnabi2022open, yuan2023support, shalabi2024ooc, zhou2022fndclip}, we utilize the powerful CLIP model because of its shared latent space, learnt from image-caption pairs. 
To effectively utilize the unlabeled data, we want to generate robust pseudo-labels which can then be used for further training the model. 
The core idea of CoVLM is to leverage additional guidance from another vision-language model BLIP, which can convert the input image into descriptive text.
This generated text, combined with the original image-text pair, is used to create robust pseudo-labels based on threshold parameters derived from the labeled data. The intuition behind determining the pseudo-labels for the unlabeled data using threshold parameters is detailed in Subsection~\ref{subsec:unlabel_data}. Subsection~\ref{subsec:extract_parameters_label_data} explains how the threshold parameters are obtained from the labeled data. Finally, Subsection~\ref{subsec:full_training} outlines the complete training procedure, incorporating both labeled and unlabeled data using the pseudo-labels.
\begin{figure}[t]
    \centering
    \includegraphics[width=1\textwidth]{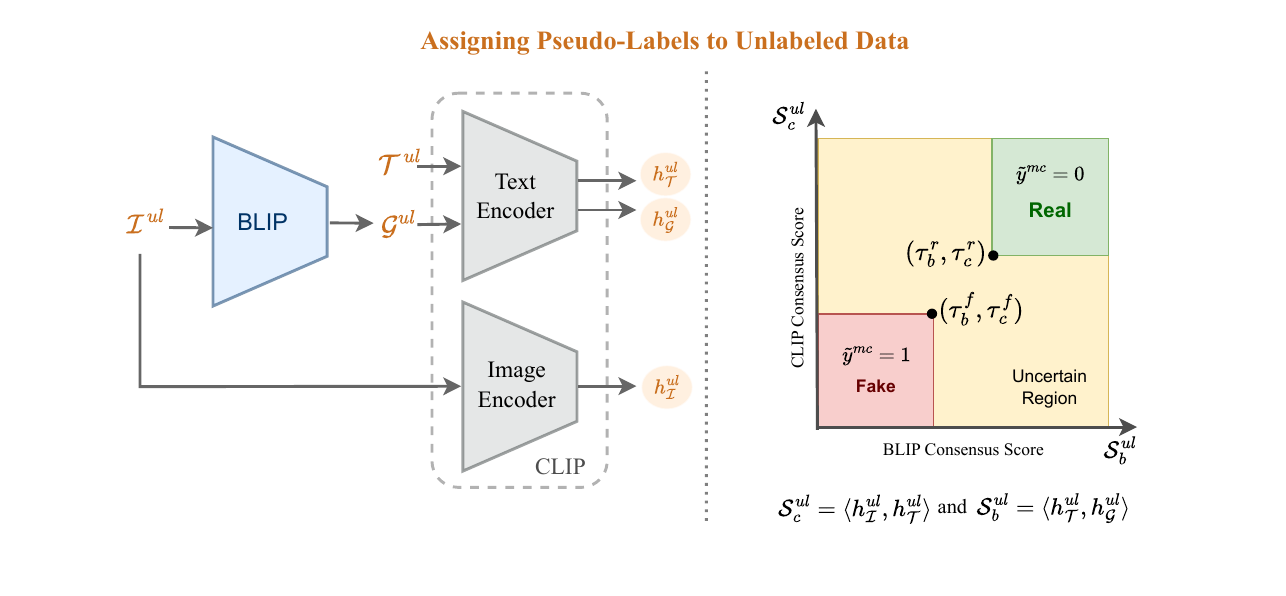}
    \caption{Illustration of the pseudo-label assignment for the unlabeled image-text pairs. Using both image and text embeddings, CLIP consensus score $\mathcal{S}_{c}$ is calculated and for given text and generated text BLIP consensus score $\mathcal{S}_{b}$ is calculated.}
    \label{fig:pseudo_label_assignment}
\end{figure}
\subsection{Unlabeled data: Pseudo-Labels using Caption Consensus}
\label{subsec:unlabel_data}
Here, we explain how we determine robust pseudo-labels for the unlabeled data in the training set.
Solely using the CLIP model to generate real/fake pseudo-labels is challenging, since realistic fake image-text pairs are semantically very close, and it is difficult to differentiate them from actual real pairs.
Towards this goal, we propose to use consensus from two vision-language models (VLMs), namely CLIP and BLIP.
For simplicity, we denote the CLIP model as $\Theta^{\text{CLIP}}$, which includes both the image and text encoders represented as $ \{\Theta_{\text{image}}^{\text{CLIP}},  \Theta_{\text{text}}^{\text{CLIP}}\}$ and the BLIP model as $\Theta^{\text{BLIP}}$. First we pass each unlabeled image $\mathcal{I}^{ul}$ through BLIP to obtain a generated caption denoted as $\mathcal{G}^{ul} = \Theta^{\text{BLIP}}(\mathcal{I}^{ul})$. 
The image and text embeddings of this unlabeled image-text pair obtained from the CLIP model is denoted as $\{ h_{\mathcal{I}}^{ul}, h_{\mathcal{T}}^{ul}\}$, by passing the image $\mathcal{I}^{ul}$  and text $\mathcal{T}^{ul}$ through the image and text encoders respectively. 
We also compute the embedding of the BLIP-generated text caption as $h_{\mathcal{G}}^{ul}  =\Theta_{\text{text}}^{\text{CLIP}}(\mathcal{G}^{ul})$. 
Now, for a real image-text pair, their corresponding embeddings in the CLIP shared latent space will be relatively closer compared to when the pair is fake. 
This is measured by the CLIP consensus score $\mathcal{S}^{ul}_{c} = \langle h^{ul}_{\mathcal{I}}, {h^{ul}_{\mathcal{T}}} \rangle$, where $\langle \cdot, \cdot \rangle$ represents the inner product between two vectors, and measures the similarity between them. 
Similarly, the BLIP consensus score, which is calculated between the embeddings of the generated text from BLIP and that of the original text as $\mathcal{S}^{ul}_{b} = \langle h^{ul}_{\mathcal{T}}, h^{ul}_{\mathcal{G}} \rangle$ is high for real pairs and low for fake pairs.
Thus, for a real image-text pair, both the scores will be high, whereas, for fake pairs, both scores will be low, as the image and text will disagree, in addition to the original and generated text being far apart.
This model consensus ensures generation of robust pseudo-labels for the unlabeled data. 
Unlabeled samples for which the models are not in agreement are not used for training at that instant. 
Note that at a later training instance, this particular image-text pair can be assigned a confident pseudo-label and can thus contribute to model training. 
The complete multi-modal consensus generated pseudo-labels is given as:
\begin{equation}
    \Tilde{y}^{mc} = 
    \begin{cases} 
    Fake & \text{if } \mathcal{S}^{ul}_{c} < \tau^{f}_{c}  \text{ and } \mathcal{S}^{ul}_{b} < \tau^{f}_{b} \\
    Real & \text{if } \mathcal{S}^{ul}_{c} > \tau^{r}_{c}  \text{ and } \mathcal{S}^{ul}_{b} > \tau^{r}_{b} \\
    Ignore &  \text{Otherwise} \\
    \end{cases}
    \label{e:y_mc}
\end{equation}
The threshold boundary parameters for fake samples $\{\tau^{f}_{c}, \tau^{f}_{b}\}$ and real samples $\{\tau^{r}_{c}, \tau^{r}_{b}\}$ play a key role in the decision-making process of pseudo-labels and are dataset dependent. 
We utilize the labeled part of the training dataset to automatically obtain these parameters.
Figure~\ref{fig:pseudo_label_assignment} shows the pseudo label assignment to the image-text pairs based the model consensus using CLIP and BLIP models. Next, we discuss how these boundary thresholds are estimated.
\begin{figure}[t]
    \centering
    \includegraphics[width=1\textwidth]{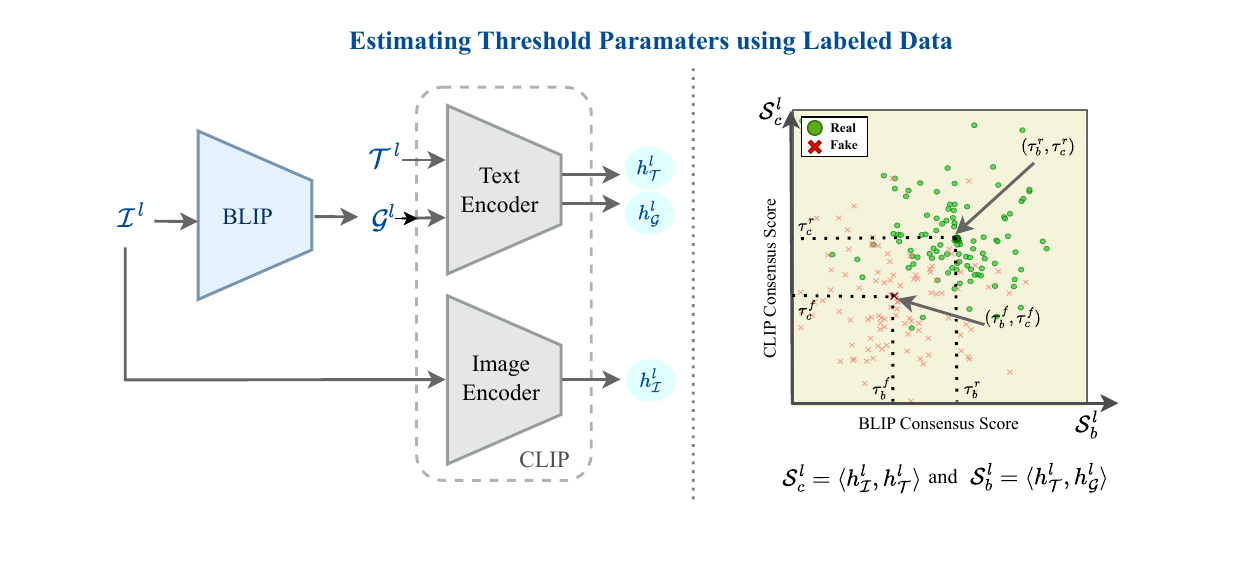} 
    \caption{Illustration of the estimation of threshold parameters using labeled data. The BLIP consensus score $\mathcal{S}_{b}$ and the CLIP consensus score $\mathcal{S}_{b}$ are calculated for all labeled samples. The mean of these labeled BLIP and CLIP consensus scores acts as threshold parameters for the unlabeled data. }
    \label{fig:gen_psuedo_label_from_label_data}
\end{figure}
\subsection{Threshold Parameters from Labeled Data}
\label{subsec:extract_parameters_label_data}
\begin{figure}[t]
    \centering
    \includegraphics[scale = 0.8]{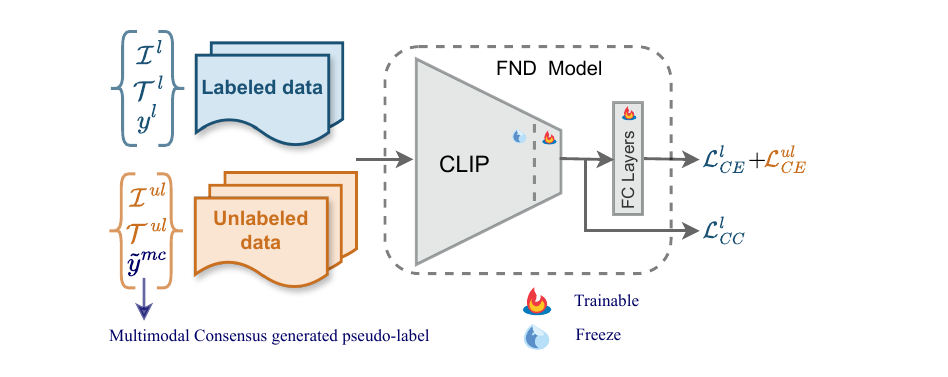} 
    \caption{Illustration of the unified training process. Both labeled and unlabeled data are used for training with binary cross-entropy loss. Contrastive clustering loss, applied to labeled data, ensures that real image-text pair embeddings are close together, while forcing fake pair embeddings far apart. This complements the pseudo-label generation process and enhances the performance of semi-supervised MFND.}
    \label{fig:unified_training}
\end{figure}

We utilize the available labeled data to automatically determine the threshold parameters, which removes the burden of manually tuning the parameters for each dataset. 
For a labeled image-text pair $(\mathcal{I}^{l}, \mathcal{T}^{l}, y^{l})$, let the BLIP generated caption be denoted as $\mathcal{G}^{l} = \Theta^{\text{BLIP}}(\mathcal{I}^{l})$. 
Now, the CLIP embeddings for the given image and text pair are given by $h_{\mathcal{I}}^{l} = \Theta_{\text{image}}^{\text{CLIP}}(\mathcal{I}^{l})$ and $h_{\mathcal{T}}^{l}  =\Theta_{\text{text}}^{\text{CLIP}}(\mathcal{T}^{l})$, and the embedding of the BLIP-generated text caption is given by $h_{\mathcal{G}}^{l}  =\Theta_{\text{text}}^{\text{CLIP}}(\mathcal{G}^{l})$. 
As shown in Figure~\ref{fig:gen_psuedo_label_from_label_data}, we calculate the consensus scores of BLIP model, $\mathcal{S}^{l}_{b} = \langle h^{l}_{\mathcal{T}}, h^{l}_{\mathcal{G}} \rangle$, and for the CLIP model, $\mathcal{S}^{l}_{c} = \langle h^{l}_{\mathcal{I}}, h^{l}_{\mathcal{T}} \rangle$ for all the labeled samples. 
Using the label $y^{l}$, we calculate the mean of these scores for both real and fake samples to obtain these threshold parameters. 
Specifically, the threshold parameters for the real class are calculated as $\tau_{b}^{r} = \sum_{i}\mathcal{S}_{b}^{i} \: \cdot \: \mathbbm{I}(y_{i}^{l} = Real)$, $\tau_{c}^{r} = \sum_{i}\mathcal{S}_{c}^{i} \: \cdot \: \mathbbm{I}(y_{i}^{l} = Real)$ (the superscript $l$ for model score $\mathcal{S}$ is removed for clarity). 
In a similar manner, we calculate the parameters for the fake class. 
Here, $\mathbbm{I}(\cdot)$ denotes the indicator function, which is 1 if the condition is true and 0 otherwise. After obtaining pseudo-labels for the unlabeled data using these parameters, the model training proceeds using both the labeled and unlabeled data, as discussed in the next section.
Figure~\ref{fig:unified_training} illustrates the unified training process of the CoVLM framework.
\subsection{Unified Training using both Labeled and Unlabeled Data}
\label{subsec:full_training}
Now, we describe the complete training process of the proposed framework for semi-supervised MFND using both the labeled and the unlabeled inputs. During training, the last few layers of CLIP image encoder $\Theta_{\text{image}}^{\text{CLIP}}$ along with two fully connected layer $\psi$ are learned. 
The predicted output for an image-text pair $(\mathcal{I}, \mathcal{T})$ is $\hat{y} = \psi(\Theta_{\text{image}}^{\text{CLIP}}(\mathcal{I}) \: \odot \: \Theta_{\text{text}}^{\text{CLIP}}(\mathcal{T}))$, where $\odot$ represents the hadamard product (element wise multiplication).  \\ \\
{\bf Learning from Labeled Data:}
For the labeled dataset $\mathcal{D}^l$, the cross-entropy loss is computed between the model's predictions and the true labels. Let $\hat{y}_i$ be the model's predicted output for the $i^{th}$ image-caption pair $(\mathcal{I}_i^l, \mathcal{T}_i^l)$, and $y_i^l$ be the true label. The binary cross-entropy loss $\mathcal{L}_{CE}^l$ is calculated by:
\begin{align} \label{e:LCE_l} 
    \mathcal{L}_{CE}^l = -\frac{1}{N_l} \sum_{i=1}^{N_l} \left( y_i^l \log \hat{y}_i + (1 - y_i^l) \log (1 - \hat{y}_i) \right)
\end{align}
Our core assumption in addressing the MFND task is that real image-text pairs will be closer in the embedding space, while fake pairs will be far apart and the pseudo-labeling for unlabeled image-text pairs also relies on this principle. 
To enforce this notion within the network, we propose to use an additional clustering objective inspired by contrastive loss given as:
\begin{align} \label{e:LCC_l} 
    \mathcal{L}_{CC}^l = -\frac{1}{N_l} \sum_{i=1}^{N_l} \left( y_i^l \log (1- \langle h^{l}_{\mathcal{I}}, h^{l}_{\mathcal{T}} \rangle)  + (1 - y_i^l) \log \langle h^{l}_{\mathcal{I}}, h^{l}_{\mathcal{T}} \rangle \right)
\end{align}
This objective encourages the real image-text pairs to come closer in the embedding space and pushes the fake pairs apart. Though this loss is on the labeled data, it helps the pseudo-labeling  process for the unlabeled data. \\ 
\begin{algorithm}[t]
\caption{Training Algorithm}
\label{alg:training}
\begin{algorithmic}[1]
\STATE \textbf{Input Model:} $\Theta^{\text{CLIP}} = (\Theta_{\text{image}}^{\text{CLIP}}, \Theta_{\text{text}}^{\text{CLIP}})$, $\Theta^{\text{BLIP}}, \psi - \text{FC Layers}$
\STATE \textbf{Training Data:}  $\mathcal{D}^{l} = \{(\mathcal{I}_i^l, \mathcal{T}_i^l, y_i^l)\}_{i=1}^{N^{l}}$, $\mathcal{D}^{ul} = \{(\mathcal{I}_i^{ul}, \mathcal{T}_i^{ul})\}_{i=1}^{N^{ul}}$ where $N^{l} \ll N^{ul}$
\FOR{each $epoch$}
    \STATE $\mathcal{B}^l$ = SampleMiniBatch($\mathcal{D}^l$) and $B^{ul}$ = SampleMiniBatch($\mathcal{D}^{ul}$) \\ 
   \textcolor[HTML]{bc5a45}{\textit{$//$ Generating BLIP captions and CLIP embeddings $//$}}
    \STATE ${\mathcal{G}}^{l} \gets \Theta^{\text{BLIP}}(\mathcal{B}^l)$ ; ${\mathcal{G}}^{ul} \gets \Theta^{\text{BLIP}}(\mathcal{B}^{ul})$
     \STATE $h^{l}_{\mathcal{I}}, h^{l}_{\mathcal{T}}, h^{l}_{\mathcal{G}} \gets \Theta^{\text{CLIP}}(\mathcal{B}^{l}, \mathcal{G}^{l})$
     \STATE $h^{ul}_{\mathcal{I}}, h^{ul}_{\mathcal{T}}, h^{ul}_{\mathcal{G}} \gets \Theta^{\text{CLIP}}(\mathcal{B}^{ul}, \mathcal{G}^{ul})$ \\ 
    \textcolor[HTML]{bc5a45}{\textit{$//$ Estimating threshold parameters from labeled data $//$}}
    \STATE $\mathcal{S}_{c}^{l} = \langle h^{l}_{\mathcal{I}}, h^{l}_{\mathcal{T}} \rangle$; $\mathcal{S}_{b}^{l} = \langle h^{l}_{\mathcal{I}}, h^{l}_{\mathcal{G}} \rangle$
    \STATE $\tau_{c}^{f}, \tau_{c}^{r}, \tau_{b}^{f}, \tau_{b}^{r} \gets \text{Estimate threshold parameters using } \mathcal{S}_{c}^{l},\: \mathcal{S}_{b}^{l} \text{ (Subsection.~\ref{subsec:extract_parameters_label_data})}$   \\
    \textcolor[HTML]{bc5a45}{\textit{$//$ Assigning pseudo-labels to unlabeled data $//$}}
    \STATE $\mathcal{S}_{c}^{ul} = \langle h^{ul}_{\mathcal{I}}, h^{ul}_{\mathcal{T}} \rangle$; $\mathcal{S}_{b}^{ul} = \langle h^{ul}_{\mathcal{I}}, h^{ul}_{\mathcal{G}} \rangle$
    \STATE $\Tilde{y}^{\text{mc}} \gets \text{Assign pseudo label using } \mathcal{S}_{c}^{ul}, \:\mathcal{S}_{b}^{ul}, \text{and estimated thresholds (Eq.~\ref{e:y_mc})}$ \\
    \textcolor[HTML]{bc5a45}{\textit{$//$ Output model predictions and loss calculations $//$}}
    \STATE $\hat{y}^{l} \gets \psi(h^{l}_{\mathcal{I}} \odot h^{l}_{\mathcal{T}}) $;  $\hat{y}^{ul} \gets \psi(h^{ul}_{\mathcal{I}} \odot h^{ul}_{\mathcal{T}}) $
    \STATE $\mathcal{L}_{CE}^{l}$ = CrossEntropyLoss(${y}^{l}$, $\hat{y}^{l}$) , $\mathcal{L}_{CC}^l$ = ContrastiveClusterLoss(${y}^{l},\mathcal{S}_{c}^{l}$)
    \STATE $\mathcal{L}_{CE}^{ul}$ = CrossEntropyLoss($\Tilde{y}^{\text{mc}}, \hat{ y}^{ul})$
    \STATE  $\mathcal{L}_{Total}$ = $\mathcal{L}_{CE}^{l}$ + $\mathcal{L}_{CE}^{ul}$ + $\lambda \mathcal{L}_{CC}^l$
\ENDFOR
\RETURN \{$\Theta^{\text{CLIP}} , \psi$\}
\end{algorithmic}
\end{algorithm}

\begin{algorithm}[t]
\caption{Inference Algorithm}
\label{alg:inference}
\begin{algorithmic}[1]
\STATE \textbf{Trained CLIP Model:} $\Theta^{\text{CLIP}} = (\Theta_{\text{image}}^{\text{CLIP}}, \Theta_{\text{text}}^{\text{CLIP}}), \psi - \text{FC Layers}$ 
\STATE \textbf{Test data:} $\mathcal{D}^{t} = {(\mathcal{I}_i^{t}, \mathcal{T}_i^{t})}_{i=1}^{N^{t}}$
\FOR {each $i \in {1, \dots, N^{t}}$}
    \STATE $\hat{y}_i \gets \psi(\Theta^{\text{CLIP}}_{\text{Image}}(\mathcal{I}_i^{t}) \odot \Theta^{\text{CLIP}}_{\text{Text}}(\mathcal{T}_i^{t}))$ 
\ENDFOR
\end{algorithmic}
\end{algorithm}
\noindent
{\bf Learning from Unlabeled Data: }
As explained earlier, we use model consensus to generate pseudo-labels for the unlabeled data, among which the confident ones satisfying the criterion in equation Eq.~\ref{e:y_mc} are used for training the model as follows:
\begin{equation} \label{e:LCE_ul} 
    \mathcal{L}_{CE}^{ul} = -\frac{1}{N^{ul}} \sum_{j=1}^{N^{ul}} \left( \Tilde{y}_j^{\text{mc}} \log \hat{y}_j + (1 - \Tilde{y}_j^{\text{mc}}) \log (1 - \hat{y}_j) \right)
\end{equation}
Here, $\hat{y}_j$ be the model's predicted output for the $j^{th}$ unlabel image-caption pair $(\mathcal{I}^{ul}_j, \mathcal{T}^{ul}_j)$ and the total loss for model training is thus given by
\begin{align*}
    \mathcal{L}_{\text{total}} = \mathcal{L}_{CE}^l + \mathcal{L}_{CE}^{ul} + \lambda \mathcal{L}^l_{CC},
\end{align*}
where $\lambda$ is a hyperparameter used to balance the contribution of the contrastive loss. The complete training procedure is summarized in Algorithm~\ref{alg:training}. After training, the BLIP model need not be stored for inference, and only the trained CLIP model is used for inference as explained in Algorithm~\ref{alg:inference}.

\section{Experiments}
\label{expts}
Here, we discuss the datasets and implementation details, followed by the experimental results.

\subsection{Dataset details}
We train and evaluate the proposed approach on three well-known multi-modal FND datasets: NewsCLIPpings~\cite{luo2021NewsCLIPpings}, GossipCop~\cite{gossipcop}, and PolitiFact~\cite{politifact}. (i) News-CLIPpings: Designed to address the evolving threat of misinformation from cheap fakes to sophisticated deep fakes, this dataset presents unmanipulated but contextually mismatched image-text pairs. (ii) GossipCop: Reflects the wide-spread nature of celebrity-related misinformation. (iii) PolitiFact: Created from news articles for fact-checking purposes. 
To the best of our knowledge, there are no existing works on SS-MFND. 
So we create the SS-MFND protocol by dividing the available datasets into a labeled and unlabeled part. 
Table~\ref{tab:table1} presents a detailed breakdown of the data samples distributed across the training, validation, and testing phases for each dataset for the semi-supervised MFND task. We also conducted experiments with different percentages of labeled and unlabeled data to test the robustness of the proposed approach.

\begin{table}[t]
\centering
\caption{Data samples distribution for training, validation and testing for semi-supervised MFND task.}
\label{tab:table1}
\begin{tabularx}{0.8\textwidth}{@{}Xcccc@{}}
\toprule
Dataset & \multicolumn{2}{c}{Training} & Validation & Testing \\
\cmidrule(lr){2-3}  
       & Labeled (5\%) & Unlabeled (95\%) &  &  \\
\midrule
NewsCLIPpings~\cite{luo2021NewsCLIPpings} & 3513 & 67555 & 7023 & 7263 \\
GossipCop~\cite{gossipcop} & 542 & 10302 & 2711 & 3388 \\
PolitiFact~\cite{politifact} & 12 & 225 & 59 & 74 \\
\bottomrule
\end{tabularx}
\end{table}

\subsection{Implementation Details}
We use CLIP-ViT/B32 \cite{radford2021learning} for image and text encoding as the backbone and the BLIP captioning large model \cite{li2022blip} for image captioning. The output of CLIP model features is followed by two fully connected layers for classification, which consists of batch normalization and dropout. Before the main training phase, both vision-language models (VLMs) are fine-tuned using the labeled data as a warm-up step. After the warm-up, we train the model for 40 epochs with a batch size of 64 across all datasets. The learning rate is set to $5 \times 10^{-4}$ over the first 20 epochs and then follows a CosineAnnealing scheduler for the remaining epochs. The Adam optimizer, combined with this learning rate schedule, is employed for the training process. All experiments are conducted on NVIDIA RTX 2080 GPUs using the PyTorch library.
\subsection{Baselines}
Since there are no existing works which have addressed the SS-MFND task, we create our own strong baselines to compare the proposed framework. 
First, we consider the supervised training of the CLIP model with the available labeled data using cross-entropy loss, referred to as Sup@5\%, which serves as the lower bound for our experiments. 
Additionally, we consider supervised training with 100\% labeled data (also referred to as Sup@100\%), representing the upper bound and the best possible performance achievable by utilizing unlabeled data and the given model. 
In addition, we also include three strong baselines using the state-of-the-art semi-supervised approaches proposed for uni-modal case, but adapted here for our task: FixMatch \cite{sohn2020fixmatch}, FreeMatch* \cite{wang2022freematch}, and Adsh \cite{guo2022class}. 
These methods leverage unlabeled data through different thresholding schemes to learn model representations. 
FixMatch employs a fixed threshold, while FreeMatch* uses learnable threshold parameters inspired by FreeMatch \cite{wang2022freematch}. Since our approach focuses on obtaining optimal threshold parameters, we denote the version with learnable threshold parameters as FreeMatch*, and Adsh generates adaptive thresholds based on class dependencies. We have extensively fine-tuned these approaches for optimal threshold parameters and report their best performance. Inspired by MFND literature \cite{luo2021NewsCLIPpings, abdelnabi2022open, shalabi2024ooc, qi2024sniffer}, we report standard test accuracy as the performance metric for comparison.

\begin{table}[t]
  \caption{Experimental results on NewsCLIPpings, GossipCop and PolitiFact datasets.}
  \label{tab:Alldatsets_Allmethods_Main_table}
  \centering
  \newcolumntype{Y}{>{\centering\arraybackslash}X}
  \newcolumntype{M}{>{\centering\arraybackslash}m{4.2cm}} 
  \begin{tabularx}{\textwidth}{ M 
    >{\hsize=.9\hsize\linewidth=\hsize\centering\arraybackslash}X 
    >{\hsize=1.2\hsize\linewidth=\hsize\centering\arraybackslash}X 
    >{\hsize=.9\hsize\linewidth=\hsize\centering\arraybackslash}X} 
    \toprule
    \textbf{Method} & \textbf{NewsCLIPpings} & \textbf{GossipCop} & \textbf{PolitiFact} \\
    \midrule
    
    Sup @ 5\% (Lower Bound) & 65.57\% & 51.91\% & 50.78\% \\ \hline
    FixMatch~\cite{sohn2020fixmatch} & 65.66\% & 73.70\% & 55.78\% \\
    FreeMatch*~\cite{wang2022freematch} & 64.06\% & 74.21\% & 59.90\% \\
    Adsh~\cite{guo2022class} & 64.78\% & 72.05\% & 50.78\% \\
    \rowcolor[HTML]{E0E0E0}\textbf{CoVLM} (Ours) & \textbf{67.34\%} & \textbf{76.42\%} & \textbf{60.00\%} \\ \hline
    Sup @ 100\% (Upper Bound) & 70.02\% & 83.76\% & 72.50\% \\
    
    \bottomrule
  \end{tabularx}
\end{table}
\begin{figure}[htb!]
    \centering
    \includegraphics[width=0.9\textwidth]{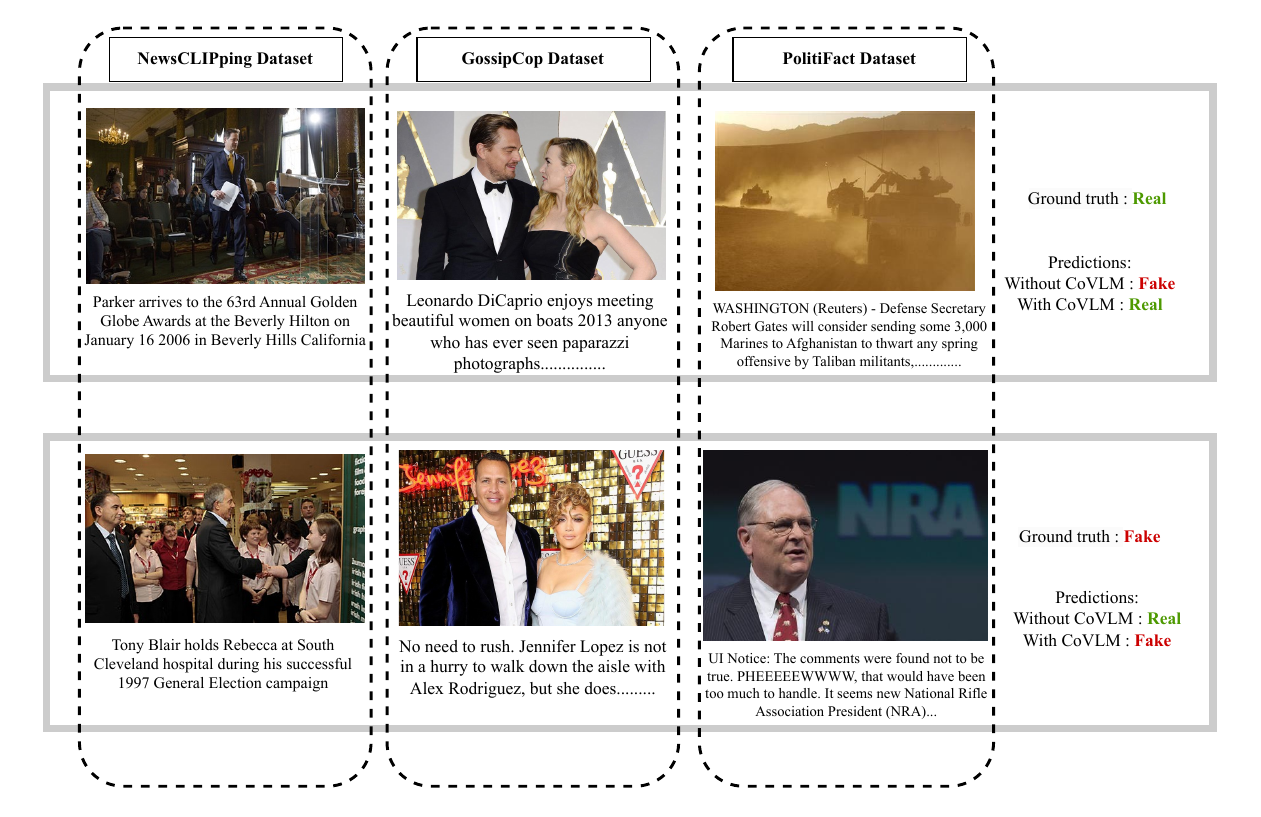}
    \caption{The first row shows real image-text pairs, which were misclassified without CoVLM. After applying CoVLM, the model correctly predicts these pairs as real. The second row shows fake image-text pairs, which CoVLM accurately identifies as fake.}
    \label{fig:visual_analysis}
\end{figure}
\subsection{Experimental Results}
Table~\ref{tab:Alldatsets_Allmethods_Main_table} reports the experimental results on NewsCLIPpings \cite{luo2021NewsCLIPpings}, GossipCop \cite{gossipcop}, and PolitiFact \cite{politifact} datasets for the semi-supervised MFND task. For NewsCLIPpings, the CLIP model trained with only 5\% labeled data (Sup@5\%) achieved 65.57\% accuracy. Utilizing the semi-supervised thresholding methods did not show any significant improvement, and in some cases even degraded the performance. 
This indicates that the thresholding methods on logit space proposed for image classification (unimodal case) may fail to capture the intricate relationship between real and fake image-text pairs. 
In contrast, the proposed CoVLM approach showed an improvement of $1.77\%$ by effectively utilizing the unlabeled data. 

For the GossipCop dataset, the baseline accuracy with only 5\% labeled data is 51.91\%. However, the proposed CoVLM significantly improves this, achieving 76.42\%. On the PolitiFact dataset, the baseline supervised accuracy is 50.78\%. Here, CoVLM demonstrates substantial improvement, achieving 60.00\%. Although there is still scope for improvement to reach the upper bound on these datasets, CoVLM significantly outperforms state-of-the-art semi-supervised approaches by a considerable margin.

Figure~\ref{fig:visual_analysis} shows a visual comparison highlighting the efficacy of the CoVLM method in identifying real and fake news across three distinct datasets: NewsCLIPpings, GossipCop, and PolitiFact. In the top row, the image-text pairs are from the Real class but were incorrectly predicted as Fake without CoVLM; with CoVLM, they were correctly predicted as Real. In contrast, the bottom row displays instances of misinformation that, without CoVLM, were wrongly classified as true stories but were correctly identified as Fake with CoVLM.

\section{Analysis and Ablation Study}
\label{analysis}
In this section, we analyze the effects of data imbalance on the MFND task, the influence of the quantity of unlabeled data in the training procedure, and the impact of each loss component of the proposed CoVLM approach.
\subsection{Impact of Data Imbalance in MFND}
As mentioned earlier, in real-world, there exists a severe data imbalance in image-text pairs for the MFND task, making it more challenging. This is primarily because the number of real image-caption pairs is usually much higher than the fake ones.
To simulate this challenging scenario, inspired from class-imbalance semi supervised learning~\cite{guo2022class}, we synthetically imbalance the NewsCLIPpings dataset with a 9:1 ratio, meaning that out of every 10 samples, 9 will be real and 1 will be fake in both labeled and unlabeled data. This imbalance makes the task more challenging and can cause the model to become biased towards predicting every sample as real, unless suitable measures are taken to handle this imbalance.  

Table~\ref{imb} shows the experimental results for this imbalanced SS-MFND protocol. 
We observe that for this scenario, methods like FixMatch and Adsh helps to improve the baseline performance to a great extent. 
But the proposed CoVLM framework significantly outperforms all the other methods, with results close to that of the balanced case.
\begin{figure*}[t]
  \centering
  
  \begin{minipage}[t]{0.4\textwidth}
  \centering
  \captionof{table}{Experiment results on imbalanced NewsCLIPpings dataset.}
    \begin{adjustbox}{max width=\linewidth}
    
    \begin{tabular}{lc}
        \toprule
        Model & NewsCLIPpings \\
        \midrule
        Sup @ 5\%                             & 51.89\% \\
        FixMatch~\cite{sohn2020fixmatch}      & 63.84\% \\
        FreeMatch*~\cite{wang2022freematch}     & 54.36\% \\
        Adsh~\cite{guo2022class}              & 63.36\% \\
        \rowcolor[HTML]{E0E0E0}\textbf{CoVLM} (Ours)  & \textbf{66.76\%} \\
        \bottomrule
    \end{tabular}
    \label{imb}
    \end{adjustbox}
    
  \end{minipage}
  \hfill
  \begin{minipage}[t]{0.55\textwidth}
  \caption{This graph illustrates the effect of amount of unlabeled Data on test accuracy on the NewsCLIPpings dataset.}
    \centering
    \begin{tikzpicture}[scale=1]
        \definecolor{color0}{HTML}{E3B448}
        \definecolor{color1}{HTML}{3A6B35}
        \definecolor{color2}{HTML}{D64161}
        \begin{groupplot}[
            group style={group size=2 by 1, horizontal sep=1.8cm},
            height=4cm, width=6cm,
            axis lines=left,
            xlabel={Availability of unlabeled data },
            ylabel={Test Accuracy (\%)},
            grid=major,
            ymin=50, ymax=80,
            xmin=0, xmax=4.5,
            xtick={0,1,2,3,4},
            xticklabels={$0\times$,$1\times$,$2\times$,$4\times$,$10\times$},
            legend style={nodes={scale=0.7, transform shape}, at={(0.45,0.95)}, anchor=north, legend columns=-1},
            xlabel style={at={(axis description cs:0.5,-0.15)}, anchor=north},
            ylabel style={at={(axis description cs:-0.15,0.5)}, anchor=south}
        ]
        \nextgroupplot[]
        \addplot[color=color0, very thick] coordinates {
            (0, 65)
            (1,66.75) 
            (2,67.69) 
            (3,67.53) 
            (4,67.68)
            
        };
        \addlegendentry{Balanced}
        \addplot[color=color2, very thick] coordinates {
            (0, 51)
            (1,62.57) 
            (2,65.99) 
            (3,66.42) 
            (4,66.76)
            
        };
        \addlegendentry{Imbalanced}
        \end{groupplot}
    \end{tikzpicture}
    
    \label{fig:graph_unlabel_times}
  \end{minipage}
\end{figure*}
\subsection{Impact of Amount of Unlabeled Data}
In real-world, the amount of unlabeled data available for training the model can vary. To analyze the proposed framework under these conditions, we evaluate its performance by varying the amounts of unlabeled data relative to the labeled data. The amount of labeled data (5\%) is fixed for all these cases, only the unlabeled data is varied. Specifically, we use the number of unlabeled data as $0\times, 1\times, 2\times, 4\times, \text{and}\: 10\times$ the number of labeled data.  From Figure~\ref{fig:graph_unlabel_times}, we observe that the model’s performance saturates at approximately $4\times$ the amount of labeled data in both balanced and imbalanced cases.

\subsection{Ablation Study}
The proposed framework is trained using three loss components. 
To demonstrate the importance of each proposed component, we conducted an ablation study on the NewsCLIPpings dataset. Table~\ref{tab:ablation_study} presents the results for both balanced and imbalanced cases. The first row represents the model trained only with the labeled cross-entropy loss ($\mathcal{L}_{CE}^l$), which essentially fine-tunes CLIP without additional guidance. Adding the contrastive clustering loss ($\mathcal{L}_{CC}^{l}$) significantly enhances the performance by enforcing better separation between real and fake pairs. Finally, incorporating the unlabeled data loss ($\mathcal{L}_{CE}^{ul}$) with robust pseudo-labels generated from the consensus of CLIP and BLIP further boosts performance, showcasing the benefits of CoVLM for semi-supervised MFND.
\begin{table}[t!]
  \centering
  \caption{Ablation study on NewsCLIPpings Dataset illustrating the importance of each component in the loss function.}
  \label{tab:ablation_study}
  \begin{tabularx}{0.8\textwidth}{@{} *{3}{>{\centering\arraybackslash}X} | *{2}{>{\centering\arraybackslash}X} @{}}
    \toprule
    $\mathcal{L}_{CE}^l$ & $\mathcal{L}_{CC}^l$ & $\mathcal{L}_{CE}^{ul}$ & Balanced & Imbalanced \\
    \midrule
    $\checkmark$ & $\times$ & $\times$ & 65.77\% & 51.89\% \\
    $\checkmark$ & $\checkmark$ & $\times$ & 66.57\% & 63.96\% \\
    \textbf{$\checkmark$} & \textbf{$\checkmark$} & \textbf{$\checkmark$} & \textbf{67.67\%} & \textbf{66.76\%} \\
    \bottomrule
  \end{tabularx}
\end{table}

\section{Conclusion}
In this paper, we introduced CoVLM, a novel framework for semi-supervised multi-modal fake news detection, designed to operate effectively with limited labeled data and a substantial amount of unlabeled data. By leveraging the consensus between two vision-language models, CLIP and BLIP, CoVLM generates robust pseudo-labels that capture the intricate relationships between images and text. Our extensive experiments on benchmark datasets such as NewsCLIPpings, GossipCop, and PolitiFact demonstrate the effectiveness  of CoVLM. Moreover, CoVLM effectively handles data imbalance, maintaining its performance even when the datasets are imbalanced to reflect real-world conditions. This work offers a significant advancement in multi-modal fake news detection, providing a practical solution for leveraging both labeled and unlabeled data.

\textbf{Acknowledgement} This work is partly supported through a research grant from SERB, Department of Science and Technology, Govt. of India (SPF/2021/000118).

\bibliographystyle{unsrt}
\bibliography{ref}
\end{document}